\documentclass[conference]{IEEEtran}
\ifCLASSOPTIONcompsoc
  \usepackage[nocompress]{cite}
\else
  \usepackage{cite}
\fi
\usepackage{amsmath,amssymb,amsfonts}
\usepackage{algorithmic}
\usepackage{graphicx}
\usepackage{textcomp}
\usepackage{xcolor}
\usepackage{hyperref}
\usepackage{float}
\def\BibTeX{{\rm B\kern-.05em{\sc i\kern-.025em b}\kern-.08em
    T\kern-.1667em\lower.7ex\hbox{E}\kern-.125emX}}

\makeatletter
\newcommand{\linebreakand}{%
  \end{@IEEEauthorhalign}
  \hfill\mbox{}\par
  \mbox{}\hfill\begin{@IEEEauthorhalign}
}
\makeatother

\begin{document}

\title{Enhancing Self-Driving Segmentation in Adverse Weather Conditions: A Dual Uncertainty-Aware Training Approach to SAM Optimization}

\author{

\IEEEauthorblockN{Dharsan Ravindran}
\IEEEauthorblockA{
    \textit{Queen's University} \\
    19dr26@queensu.ca
}

\and
\IEEEauthorblockN{Kevin Wang}
\IEEEauthorblockA{
    \textit{Queen's University} \\
    22wsl3@queensu.ca
}

\and
\IEEEauthorblockN{Zhuoyuan Cao}
\IEEEauthorblockA{
    \textit{Queen's University} \\
    23xl1@queensu.ca
}


\linebreakand 
\IEEEauthorblockN{Saleh Abdelrahman}
\IEEEauthorblockA{
    \textit{Queen's University} \\
    22qfc1@queensu.ca
}
\and
\IEEEauthorblockN{Jeffery Wu}
\IEEEauthorblockA{
    \textit{Queen's University} \\
    23xcqg@queensu.ca
}

} 
\maketitle

\begin{abstract}
Recent advancements in vision foundation models like Segment Anything Model (SAM) and its successor SAM2 have established new state-of-the-art benchmarks for image segmentation tasks. However, these models often fail in inclement weather scenarios where visual ambiguity is prevalent, primarily due to their lack of uncertainty quantification capabilities. Drawing inspiration from recent successes in medical imaging—where uncertainty-aware training has shown considerable promise in handling ambiguous cases. We explore two approaches to enhance segmentation performance in adverse driving conditions. First, we implement a multi-step finetuning process for SAM2 that incorporates uncertainty metrics directly into the loss function (1) to improve overall scene recognition. Second, we adapt the Uncertainty-Aware Adapter (UAT) originally developed for medical image segmentation (2) to autonomous driving contexts. We evaluate these approaches on three diverse datasets: CamVid(1,2), BDD100K(1), and GTA driving(1). Our experimental results demonstrate that UAT-SAM outperforms standard SAM in extreme weather scenarios, while the finetuned SAM2 with uncertainty-aware loss shows improved performance across overall driving scenes. These findings highlight the importance of explicit uncertainty modeling in safety-critical autonomous driving applications, particularly when operating in challenging environmental conditions.
\end{abstract}

\section{\textbf{Introduction}}

Inclement weather poses significant hurdles for image perception in self-driving systems, as cameras are critical for tasks like object detection, lane recognition, and traffic sign interpretation, which rely heavily on clear visual data\cite{ZHANG2023146}. Adverse conditions such as rain, snow, fog, or sleet degrade image quality through raindrop-obscured lenses, snow accumulation, fog-induced contrast loss, or glare from wet surfaces, introducing noise and distortion that confuse computer vision algorithms. To address this, researchers are exploring techniques like real-time image enhancement using convolutional neural networks (CNNs) or generative adversarial networks (GANs) to "clean" raw camera feeds, alongside training models on synthetic or augmented datasets that simulate weather-corrupted visuals\cite{sining}. However, dynamic or extreme conditions still challenge these methods.

Current state-of-the-art self-driving systems are black-box ML models that provide little insight to their decision making process. When encountering uncertain conditions, such machines can give outputs which cause actions that put those in and around the vehicles at risk. Especially in high-risk and high-volatility scenarios where lives and bodies may be at stakes, making safe decisions requires large amounts of certainty on the accuracy of the information it uses. One way to address this concern has been the introduction of uncertainty quantification, a way for models to give a clear sign of how confident they are in the results they are outputting. Using this metric, users can better understand when to use the given results and when their models struggle.

Both  of the aforementioned problems are exacerbated by the inherent uncertainty introduced by adverse weather, which is rarely quantified or leveraged effectively in existing approaches. Without explicit modeling of this uncertainty, segmentation models cannot appropriately adapt their confidence levels or focus computational resources on the most challenging regions. This limitation is particularly problematic in the context of foundation models like SAM and SAM2, which, despite their impressive capabilities in standard conditions, lack specific mechanisms to handle the uncertainty introduced by inclement weather.

Our research addresses these challenges through two complementary uncertainty-aware approaches: one targeting the extreme conditions where object detection becomes critical for safety, and another improving overall segmentation quality across varying weather conditions. These approaches seek to enhance the robustness of autonomous driving perception systems by explicitly incorporating uncertainty estimation into the segmentation process, thereby enabling more reliable operation in the dynamic and unpredictable environmental conditions encountered in real-world driving scenarios.

\section{\textbf{Related Work}} \label{relatedwork}

Recent advancements in computer vision have led to significant improvements in semantic segmentation models, particularly with the introduction of the Segment Anything Model (SAM) \cite{kirillov2023segment}. SAM represents a paradigm shift in segmentation approaches, utilizing a prompt-based architecture that enables zero-shot segmentation across diverse domains. Building upon this foundation, SAM2 \cite{ravi2024sam2} further enhances these capabilities with improved performance and efficiency. These models have demonstrated remarkable versatility across various applications \cite{yang2024samurai,chen2023semanticSA} but face challenges in complex environmental conditions such as those encountered in autonomous driving scenarios.

Uncertainty estimation in deep learning has emerged as a critical research direction\cite{dutta2023estimating}, particularly for safety-critical applications like autonomous driving. The seminal work by \cite{kendall2017} established a framework for distinguishing between epistemic uncertainty (model uncertainty) and aleatoric uncertainty (data uncertainty), both of which are essential for reliable decision-making systems. Similar implementations demontrated how techinques such as Monte Carlo dropout could provide practical approximations of Bayesian inference in deep neural networks, offering computationally efficient uncertainty estimates \cite{dawood2023ENN&UAT}. These approaches have since been extended to various computer vision tasks, including semantic segmentation.

In the medical imaging domain, uncertainty-aware training has proven particularly valuable, with several studies demonstrating improved segmentation performance in regions with ambiguous boundaries or pathological variations. Many SAM models like SAM-Med2D, have successfully improved CT and MRI image segmentation by finetuning adapters in the SAM architecture\cite{cheng2023sammed2d}. Despite their improvements medical images are often ambiguous. Physicians often provide different annotations for lesions in CT images \cite{jiang2024UATadapt}.

The seminal paper on Uncertainty-Aware Adapter: Adapting Segment Anything Model (SAM) for Ambiguous Medical Image Segmentation by \cite{jiang2024UATadapt} provides a strong foundation for addressing perceptual ambiguity through aleatoric uncertainty modeling. This architecture creates a dedicated latent space for sampling possible segmentation variants, building upon previous uncertainty works like Probabilistic U-Net \cite{kohl2019s}. Its core innovation is the Condition Modifies Sample Module (CMSM), which establishes a deeper integration between uncertainty samples and model features, unlike previous approaches that simply concatenate stochastic samples at the output layer. The Uncertainty-Aware Adapter serves as a lightweight component that can be attached to the pre-trained SAM model, preserving SAM's powerful foundation while enabling the generation of multiple plausible segmentation hypotheses. Rather than relying on one-to-one ground truth to image mappings, it calibrates the model on real-world scenarios with multiple valid interpretations. 

This approach mirrors the challenges faced in autonomous driving during inclement weather, where environmental conditions create similar perceptual ambiguities\cite{burnett2023boreas}. Just as medical images contain regions where multiple expert interpretations are valid, driving scenes during snow, rain, or fog present objects with unclear boundaries and varying visibility. The ability to generate multiple plausible segmentation hypotheses rather than a single prediction enables more robust decision-making in safety-critical autonomous systems, allowing for conservative action planning when uncertainty is high. These medical applications provide valuable insights that can be transferred to autonomous driving, particularly in identifying critical regions under adverse conditions where traditional deterministic segmentation approaches may fail due to reduced sensor reliability.

Our work leverages three prominent datasets to validate our approach: CamVid \cite{camvid}, BDD100K\cite{yubdd100k}, and GTA driving. CamVid has established itself as one of the most popular benchmarks for evaluating semantic segmentation in driving scenarios, providing high-definition video sequences with pixel-level annotations. BDD100K, developed by Berkeley, offers diverse driving scenes across different weather conditions and times of day. The GTA driving dataset complements these with synthetic driving scenes and perfect ground truth annotations. While several studies have utilized these datasets to evaluate segmentation algorithms across varying conditions, comprehensive analysis of uncertainty estimation remains limited\cite{Modas_2020, wang}. Together, these three datasets provide a rich foundation for evaluating uncertainty-aware segmentation approaches in autonomous driving applications, particularly for addressing challenges posed by inclement weather.

While existing research has made significant progress in both uncertainty estimation and robust segmentation for autonomous driving, there remains a gap in effectively combining these approaches to address the specific challenges posed by inclement weather. This paper builds upon prior work by integrating uncertainty-aware training techniques with state-of-the-art segmentation models (SAM and SAM2) to develop complementary approaches that address different aspects of the inclement weather challenge: one aimed at improving overall accuracy through uncertainty-guided finetuning \textbf{(SAM2 with Multistep Finetuning for Overall Accuracy
Improvement)}, and another focused on extreme conditions through adaptive region focusing \textbf{(UAT Adapter with SAM for Extreme Weather Conditions)}.

\section{Methodology}

\subsection{\textbf{SAM2 with Multistep Finetuning for Overall Accuracy Improvement}}
\begin{figure*} 
\centering
\includegraphics[width=\textwidth]{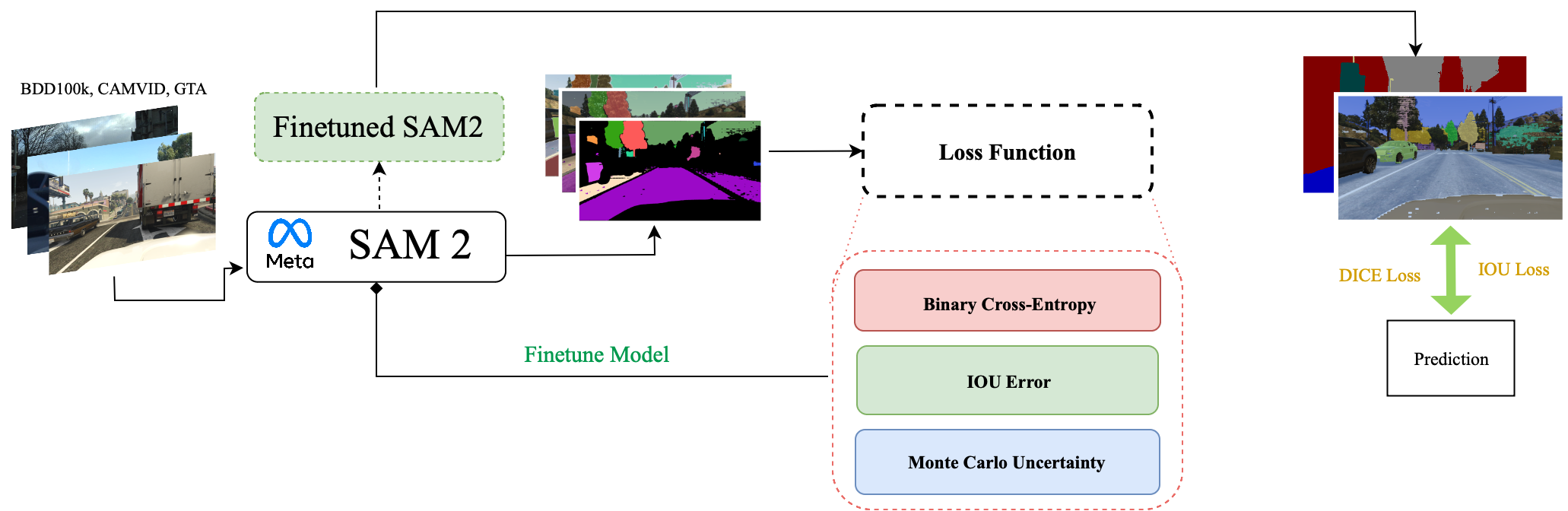}
\caption{Overview of Finetuned SAM2 architecture, along with loss function implementations, and inference testing steps.}
\label{cnn}
\end{figure*}


The following outlines our step-by-step approach for fine-tuning SAM2 on driving datasets like Bdd100k using a custom loss function. We aim to improve segmentation accuracy by incorporating multiple loss components, such as Binary Cross-Entropy, Intersection-over-Union (IoU) Error, and Monte Carlo Uncertainty Loss.

We first prepare the dataset by pairing each image with its corresponding ground truth segmentation mask for ease-of-access during training. Additionally, the ground truth masks are separated into a list of binary masks based on color, where a pixel has a value of $1$ if it is the member of the $i$'th mask and 0 otherwise.

Then, the preprocessed images are passed through SAM2 for the output masks, which have shape $[n,h,w,c]$ ($n$ masks, each is $h \times w$ pixels with $c$ channels). This is reduced to $[n,h,w]$ using sigmoid before feeding into the loss function to avoid dimension conflicts.

Once the model has made its predictions, we feed the predictions and ground truth into our custom loss function. This function incorporates three key components: 
\begin{itemize}
    \item Binary Cross-Entropy Loss
    \begin{itemize}
        \item This is used so that the model learns any underlying distributions found in driving images. 
        \end{itemize}
        $$\text{BCE}(\hat{y},y)=-\frac{1}{N}\sum_{i=1}^Ny_i \cdot \log(\sigma(y_i)) + (1-y_i) \cdot \log(1-\sigma(\hat{y}_i))$$ where \begin{itemize}
            \item $\hat{y}$ contains the predicted masks, as logits
            \item  $\vec{y}$ contains the ground truth masks
            \item $\sigma$ is the sigmoid function
            \item  $N$ is the total number of masks output by SAM2.
    \end{itemize}
    \item IoU Loss
    \begin{itemize}
        \item This is used so that the model is penalized for incorrect segmentations, such as missing a part of a truck or over-segmenting several different objects as one in its output. 
    \end{itemize}
    $$\text{IoU\_Loss}(\hat{y},y)=1-\text{IoU}(\hat{y},y)$$
    $$\text{IoU}(\hat{y},y)=\frac{\sum(\hat{y} \cdot y)+\epsilon}{\sum \hat{y} + \sum y - \sum (\hat{y} \cdot y) + \epsilon}$$ where \begin{itemize}
        \item $\hat{y}$ is the predicted mask after applying the sigmoid function
        \item $y$ is the ground truth mask
        \item $\epsilon$ is a small constant used for numerical stability.
    \end{itemize}

    \item Monte Carlo Uncertainty Loss 
    \begin{itemize}
        \item The input image is fed into the model 10 times to produce 10 mask predictions. We then calculate pixel-wise standard deviation of each pixel's assigned mask, producing a tensor of float values. 
        \item We use this tensor to as weights in our final loss function as doing so will direct the model's attention towards reducing uncertainty and variability in output.
        \item We will refer to this tensor as $U$ with width $w$m height $h$, and elements $u_{ij}$.
    \end{itemize}
    Let: $$C=\alpha \cdot \text{BCE}(\hat{y},y)+(1-\alpha) \cdot \text{IoU}(\hat{y},y)$$ $$W=C \cdot \exp(-U)$$, and $$R=\beta \cdot \frac{1}{w\times h}\sum^w_{i=1}\sum^h_{j=1}u_{ij}$$

    The combined loss function used is $$\text{Loss}=mean(W)+R$$
    
\end{itemize}

After this loss is calculated, it is backpropagated into SAM2. This updates the model's weights for improved accuracy of segmentation masks and is essential to finetuning the model for aligning its predictions with the ground truth of the dataset.

To ensure convergence, we repeat the above for 6000 steps, progressively refining the model's ability to accurately segment objects in self-driving scenarios.

The SAM2 model used was derived from the official Facebook Research's implementation \cite{ravi2024sam2}. The model is fine-tuned using Python, specifically the Pytorch framework, and is trained using a NVIDIA GeForce 4060 GPU.

\subsection{\textbf{UAT Adapter with SAM for Extreme Weather Conditions}}

Our second complementary approach was to tackle specific object instance segmentations in extreme weather scenarios utilizing the UAT-SAM adapter architecture by \cite{jiang2024UATadapt}. As referenced in section \ref{relatedwork}, the UAT adapter is a novel addition to the original SAM architecture, inspired by methodologies in medical imaging. This adapter is inserted into each transformer block of SAM. It acts as a compact set of parameters that incorporates additional information—in this case, uncertainty. The UAT adapter utilizes the CMSM (Condition Modifies Sample Module) to incorporate a sampled uncertainty code, z, derived from a CVAE (Conditional Variational Autoencoder). This CVAE employs both a Prior Net (P) and a Posterior Net (Q) to encode observed uncertainty information from the input image. \cite{jiang2024UATadapt}

Unlike previous approaches that directly concatenate the sampled code z with the main features, the UAT adapter takes a more refined approach. It integrates position vectors (p) and employs learnable attention-like mechanisms to transform z into meaningful features. These features are then combined with the main features in a layer-specific manner, allowing for nuanced modifications. This design ensures that the uncertainty sample from the CVAE is effectively captured and utilized, leading to more robust segmentation outputs.

Prior to training the CAMVID dataset required extensive pre-processing to be utilized. We applied a random weather filter either fog, rain, or snow in random filtering strengths form 0-1 (0=clear image , 1= completely obscured image) to the original images to introduce difficulty to the model when training and  testing on obscured images due to extreme weather.

Due to the module's architecture and basis on medical image segmentation, it required multiple ground truth segmentations for every image. However a majority of publicly available driving datasets including CAMVID only provide 1 set of ground truth masks. 
To simulate the ambiguous segmentation requirements in our training data, we applied elastic deformations to all 1,419 human-segmented ground truth segmentations from the CAMVID dataset. Each segmentation was deformed by randomly shifting pixel locations in both the x and y directions using a Gaussian filter. The magnitude and smoothness of these shifts were controlled by two parameters: alpha and sigma. The parameter alpha controlled the strength of the deformation, while sigma determined the smoothness of the deformation, with higher values resulting in more gradual, blurry shifts. To introduce variability and simulate different environmental conditions, three different parameter sets were used: Fog-like deformations with alpha = 20.0 and sigma = 15.0 for smoother, more blurred boundaries; Rain-like deformations with alpha = 25.0 and sigma = 4.0 for sharper, more localized changes; and Snow-like deformations with alpha = 30.0 and sigma = 7.0 for medium smoothness with stronger distortion. 
\begin{figure}[H]
\centering
\includegraphics[width=\columnwidth]{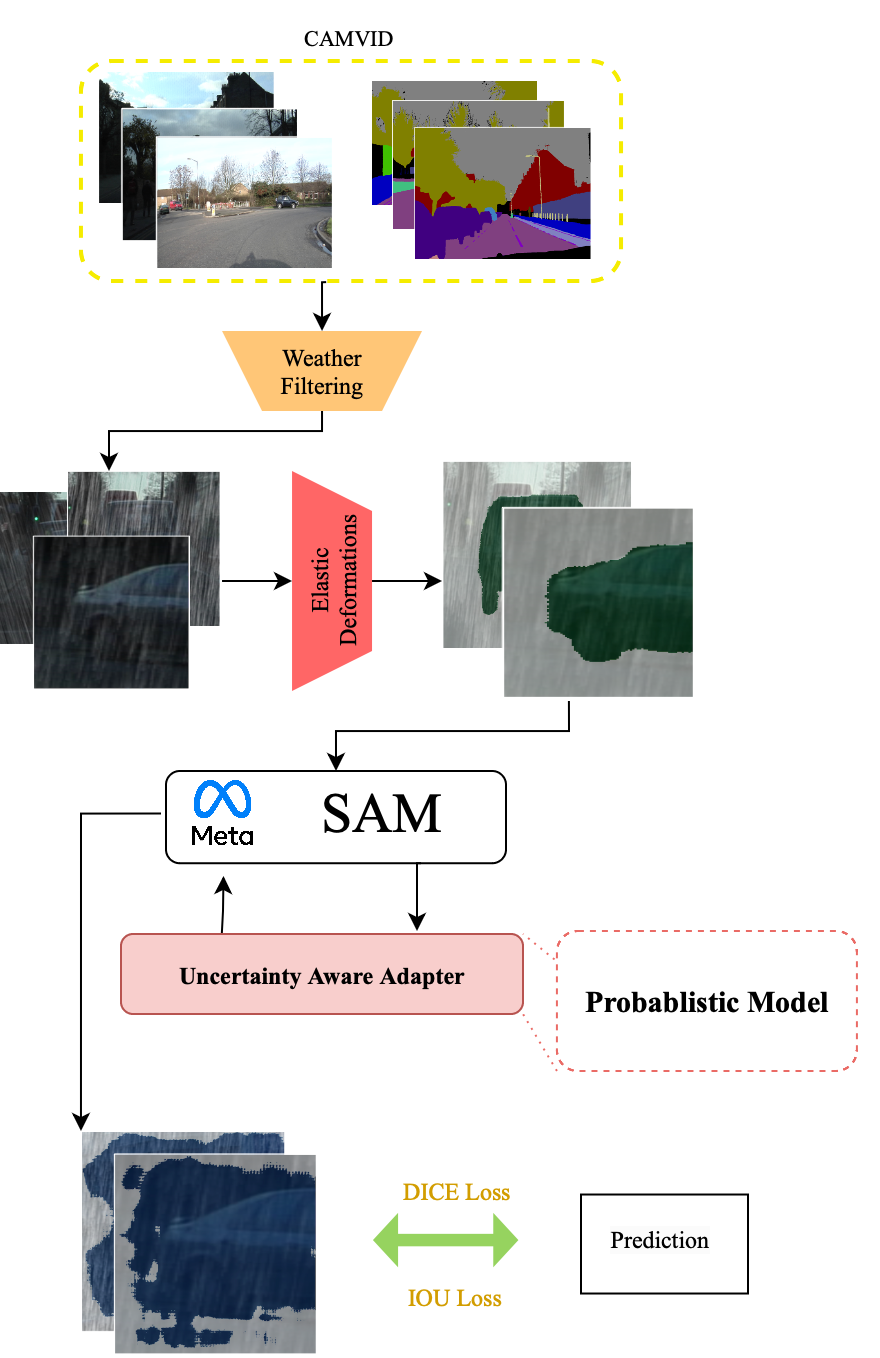}
\caption{Overview of UA-SAM Training and data augmentation pipeline.}
\label{fig:image2}
\end{figure}
This approach was used to generate three additional annotations from single ground truth, alongside one human annotation. Each image in the dataset, therefore, has four segmentation masks including ground truth, capturing a range of plausible interpretations. Each of the segmentations were matched to the original weather filtered images for training.

\begin{table*}[h]
\centering
\caption{Average IoU Scores of Multistep Finetuned SAM2}
\begin{tabular}{|l|c|c|c|c|c|c|c|c|}
\hline
\multicolumn{1}{|c|}{\textbf{Model}} & \textbf{Car} & \textbf{Truck} & \textbf{Person} & \textbf{Bicycle} & \textbf{Motorcycle} & \textbf{Traffic Light} & \textbf{Stop Sign} & \textbf{Fire Hydrant} \\ \hline
Finetuned SAM2 & 0.156 & 0.187 & 0.230 & 0.247 & 0.117 & 0.300 & 0.185 & 0.119 \\ \hline
Zero-shot SAM2 & 0.087 & 0.110 & 0.188 & 0.155 & 0.070 & 0.200 & 0.230 & 0.119 \\ \hline
\end{tabular}
\end{table*}

\begin{table*}[h]
\centering
\caption{Average DICE Coefficient Scores of Multistep Finetuned SAM2}

\begin{tabular}{|l|c|c|c|c|c|c|c|c|}
\hline
\textbf{Model} & \textbf{Car} & \textbf{Truck} & \textbf{Person} & \textbf{Bicycle} & \textbf{Motorcycle} & \textbf{Traffic Light} & \textbf{Stop Sign} & \textbf{Fire Hydrant} \\ \hline
Finetuned SAM2 & 0.333 & 0.406 & 0.531 & 0.565 & 0.259 & 0.672 & 0.416 & 0.239 \\ \hline
Zero-shot SAM2 & 0.142 & 0.200 & 0.428 & 0.333 & 0.138 & 0.475 & 0.543 & 0.239 \\ \hline
\end{tabular}

\end{table*}
We also utilized instance cropping on the data to specifically focus on car segmentations to train, leveraging uncertainty modeling to prioritize regions with high variability. This adaptation allows the model to generate accurate outputs even with noisy or ambiguous data.
The training methodology incorporates a tailored loss function, primarily the Dice Coefficient Loss, to handle segmentation. This loss function improves boundary detection, crucial for imbalanced datasets and difficult scenarios.
The training pipeline follows a multi-stage process, starting with a pre-trained Segment Anything Model (SAM) and selective parameter freezing to retain SAM’s pre-trained capabilities. Gradual adaptation fine-tunes the model for specific domain needs. Key metrics like Dice score and Intersection over Union (IoU) are monitored, with early stopping to prevent overfitting. TensorBoard visualizes the training process, ensuring high performance and adaptability in severe conditions. We tested the finetuned model and zero-shot sam by running inference on 177 heavy weather filtered CAMVID car instance segmentations with the original ground truth segmentation paired, and compared IOU and DICE across both.

\section{\textbf{Results and Discussion}}

\subsection{\textbf{SAM2 Multistep Finetuning}}

\subsubsection{Overall Accuracy Improvements}

To assess the performance of finetuned SAM2, we evaluated it against base SAM2 using Intersection-Over-Union (IoU) and the DICE coefficient. Specifically, we found the average IoU scored by the models when segmenting commonly seen road objects (e.g. cars, people, bicycles, traffic lights) as well as when segmenting entire images across the Bdd100k and Camvid datasets. The following tables describe the results we found.

\begin{table}[H]
\centering
\caption{Average IoU Scores of Multistep Finetuned SAM2}
\begin{tabular}{|l|c|c|}
\hline
\multicolumn{1}{|c|}{\textbf{Model}} & \textbf{Overall - Bdd100k} & \textbf{Overall - Camvid}  \\ \hline
Finetuned SAM2                                  & 0.303 &  0.303   \\ \hline
Zero-shot SAM2                                  & 0.246 &  0.246    \\ \hline
\end{tabular}

\end{table}

\begin{table}[H]
\centering
\caption{Average DICE Coefficient Scores of Multistep Finetuned SAM2}
\begin{tabular}{|l|c|c|}
\hline
\multicolumn{1}{|c|}{\textbf{Model}} & \textbf{Overall - Bdd100k} & \textbf{Overall - Camvid}  \\ \hline
Finetuned SAM2                                  & 0.690 &  0.690   \\ \hline
Zero-shot SAM2                                  & 0.550 &  0.550    \\ \hline
\end{tabular}

\end{table}


Our fine-tuned SAM2 outperformed zero-shot SAM2 in most classes based on IoU and DICE scores, except for stop signs and fire hydrants. This may have been the result of a class imbalance in the Bdd100k dataset, which likely contains more examples of common road objects, such as cars, people, and motorcycles, than lesser seen objects, such as stop signs and fire hydrants. Additionally, the smaller size of stop signs and fire hydrants may have contributed to the reduced segmentation performance, especially when attempting to segment them at a distance.
On average, our fine-tuned SAM2 model improved IoU by 36.13\% over zero-shot SAM2, with the highest gain being in car segmentation (+79.13\%) and the smallest nonzero gain in person segmentation (22.34\%). For DICE scores, our model improved by 48.79\% on average, with cars showing the highest increase (+134.51\%) and people showing the smallest nonzero increase (+24.07\%).

\subsubsection{Uncertainty-Aware Finetuning Benefits:}


\begin{figure}[H]
\centering
\begin{minipage}{0.48\linewidth}
    \centering
    \includegraphics[width=\linewidth]{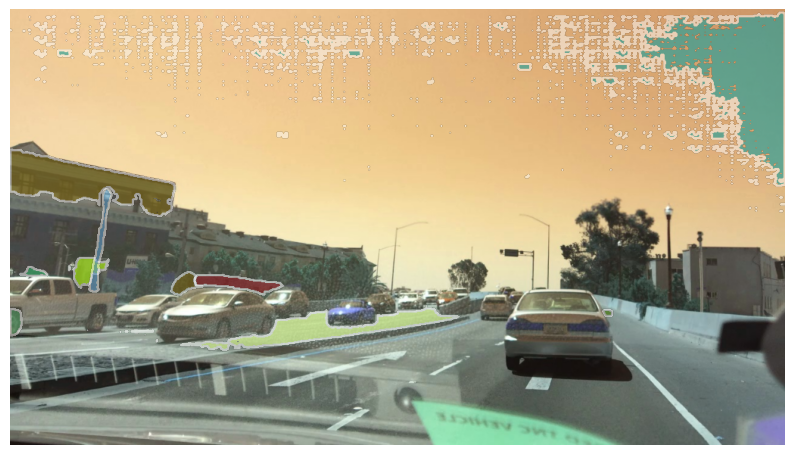} 
\end{minipage}
\hfill
\begin{minipage}{0.48\linewidth}
    \centering
    \includegraphics[width=\linewidth]{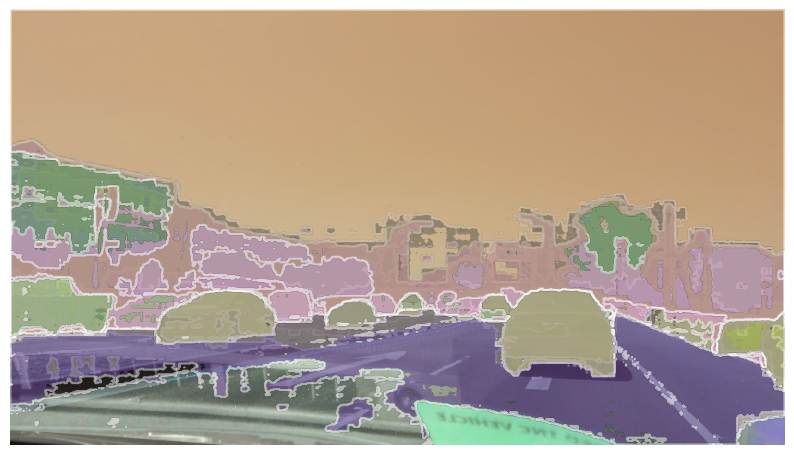} 
\end{minipage}
\caption{Performance Before and After Finetuning. Note how zero-shot SAM2 fails to generate meaningful masks over most of the image.}
\label{fig:before_after}
\end{figure}

Incorporating uncertainty into our finetuning process inmproved segmentation in ambiguous regions, particularly for multi-component objects like vehicles. Zero-shot SAM2 often produced inconsistent masks for vehicles, segmenting individual components (such as wheels or windows) or omitting the vehicle entirely (fig.  \ref{fig:before_after}). After applying uncertainty-aware finetuning, SAM2 consistently assigned a single mask per vehicle (fig. \ref{fig:before_after}), enhancing segmentation accuracy and reducing fragmented outputs.

Additionally, our finetuned SAM2 model demonstrated strong generalization across diverse driving scenarios. We evaluated its performance on datasets from various environments, including Bdd100k (recorded in New York, San Francisco, and other regions), CamVid (recorded in Cambridge), and the GTA5 Driving Dataset (recorded in a simulated driving environment). Across all three datasets, our model consistently segmented key classes such as cars, trucks, roads, and pedestrians, highlighting its robustness in both real world and synthetic driving conditions.


\subsection{\textbf{UA-SAM}}
The fine tuned UAT-Adapter SAM was tested on 177 heavy weather filtered CAMVID car instance images with the original human segmentations serving as the ground truth. SAM served as the baseline, and once again the evaluation metrics were IOU and the DICE coefficient. When heavily obscured SAM often failed to segment or contour any object within the image and was halted  by rain, and snow specifically. 

Table V demonstrates the improvements that UAT Adapter SAM was able to make over zero-shot SAM in similar scenarios, highlighting its enhanced ability to handle complex segmentation tasks. UAT-SAM showed a ~30\% increase in the DICE coefficient and ~42.7\% increase in IOU scores. The UAT Adapter SAM consistently outperformed the zero-shot SAM by focusing on regions with high variability, improving segmentation accuracy in challenging environments where visibility is compromised.

\begin{table}[H] 
\centering
\caption{AVERAGE IOU \& DICE SCORES OF UA-SAM}
\begin{tabular}{|l|c|c|}
\hline
\multicolumn{1}{|c|}{\textbf{Model}} & \textbf{Dice Score} & \textbf{IOU}  \\ \hline
Zero-shot SAM                                  & 0.4809           &  0.3221   \\ \hline
UA-SAM                                  & 0.6258           & 0.4598    \\ \hline

\end{tabular}
\end{table}

\begin{figure}[H]
\centering
\includegraphics[width=\columnwidth]{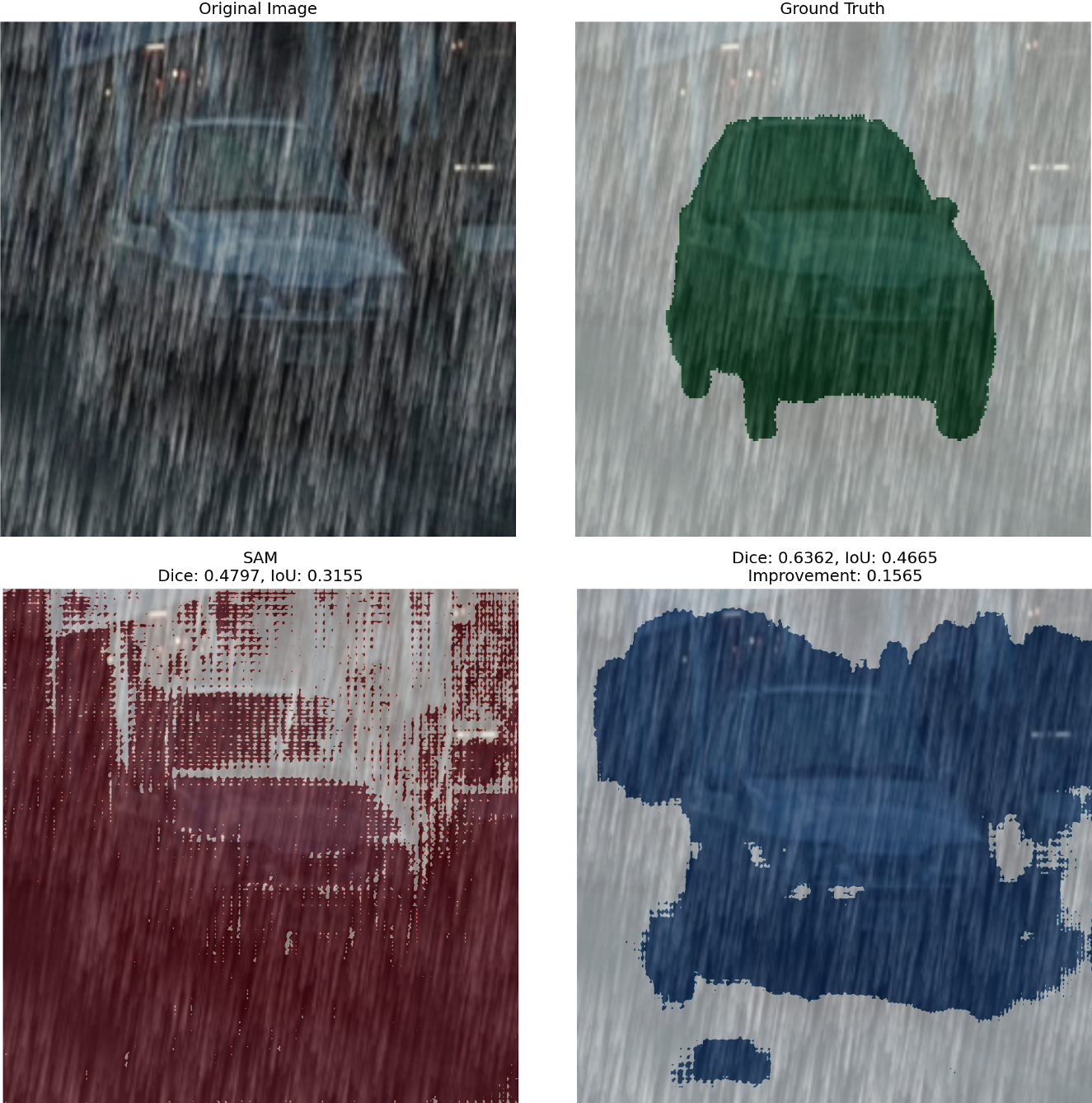}
\caption{Example of instance segmentation on heavilty filtered car image in rain scenario. From Left to Right, Top to Bottom: Filtered Original Image, Elastic Deformation GT Segmeantation, Base SAM Segmentation, UA-SAM Segmentation}
\label{fig:image4}
\end{figure}

Figure \ref{fig:image4} shows an instance of a heavily filtered image, the ground truth segmentation and the lack of any segmentations on base SAM. Although UAT-SAM can be overconfident it still is able to generally localize the car.

\begin{figure}[H]
\centering
\includegraphics[width=\columnwidth]{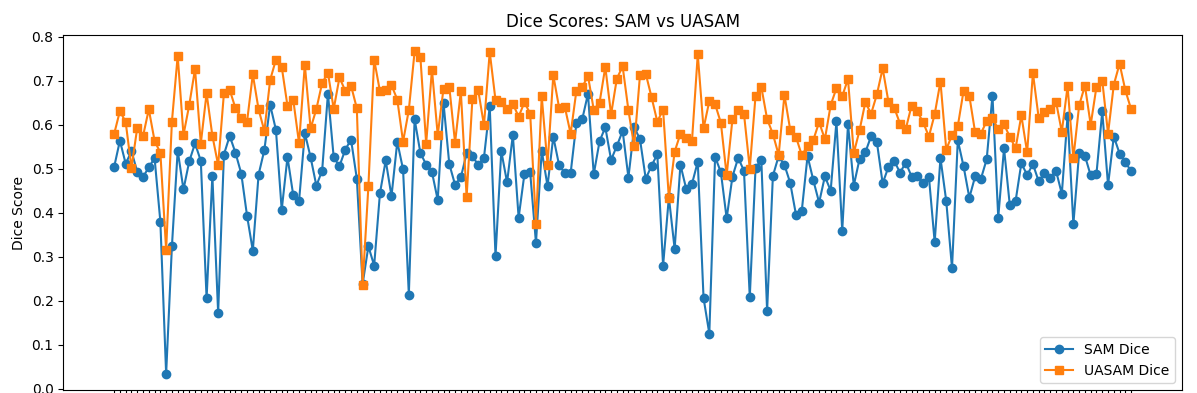}
\caption{DICE scores of Zero-shot SAM and UA-SAM across 177 car object patch segmentation in inclement weather}
\label{fig:image5}
\end{figure}

\begin{figure}[H]
\centering
\includegraphics[width=\columnwidth]{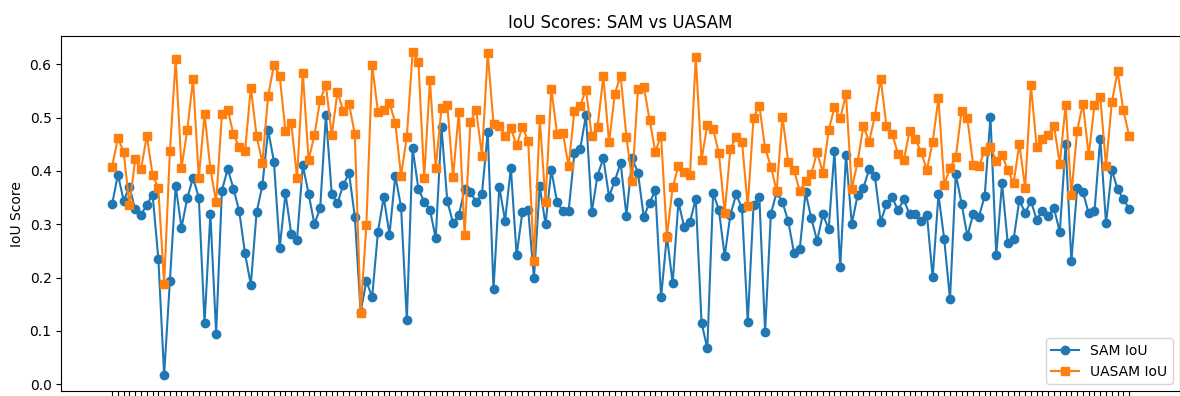}
\caption{IOU scores of Zero-shot SAM and UA-SAM across 177 car object patch segmentation in inclement weather}
\label{fig:image6}
\end{figure}

Despite Figures \ref{fig:image5} and \ref{fig:image6} demonstrating certain instance segmentations where UAT Adapter SAM and zero-shot SAM fail to segment effectively, generally, the UAT Adapter SAM exhibits better robustness and handles more challenging segmentation cases. These failures, while notable, are less frequent and often occur in particularly ambiguous or noisy regions, which further emphasizes the model’s strength in most typical conditions.
Utilizing the approach outlined by \cite{jiang2024UATadapt} to segment ambiguous segmentations in driving scenarios due to inclement weather showed considerable promise. This method enabled the model to focus on areas of uncertainty to segment cars, allowing for more reliable segmentation even under adverse weather conditions like fog, rain, or snow.

\section{Conclusion}
This research has demonstrated the effectiveness of two complementary uncertainty-aware approaches for improving semantic segmentation in self-driving applications, particularly under challenging weather conditions. The UAT adapter integrated with SAM successfully enhanced segmentation capabilities in severe weather scenarios by leveraging uncertainty estimates to identify and focus on critical regions where visibility is compromised. Our experiments on the BDD100K and CamVid datasets revealed that this approach significantly improved detection and segmentation of crucial road elements, with a particular focus on vehicles and specific objects of interest. The UAT adapter showed remarkable improvements in car detection accuracy under fog, heavy rain, and low-light conditions, where traditional segmentation methods typically fail.

In contrast, the uncertainty-incorporated multistep finetuning approach with SAM2 proved particularly effective at improving overall scene segmentation quality across varying weather conditions. This method delivered clearer contours and better distinction between foreground and background elements, resulting in more precise boundary delineation and improved class separation. The uncertainty-guided loss function enabled the model to adaptively focus on ambiguous regions during training, leading to more reliable segmentations with well-calibrated confidence estimates.

Together, these approaches address different but complementary aspects of the inclement weather challenge in autonomous driving perception. The UAT adapter provides a targeted solution for the most severe conditions where safety-critical decisions must be made despite limited visibility, while the uncertainty-finetuned SAM2 offers broader improvements in segmentation quality that enhance overall system performance.

Our contributions not only advance the state of the art in semantic segmentation for challenging conditions but also demonstrate the value of incorporating uncertainty awareness into modern foundation models like SAM and SAM2. The methods presented here have potential applications beyond autonomous driving, particularly in other safety-critical domains where perception systems must operate reliably despite environmental challenges.

\section{Future Work}
We plan to extend the UAT adapter's capabilities to segment a wider range of objects beyond cars, including pedestrians, cyclists, traffic signs, and other road users, providing a more comprehensive perception system for autonomous vehicles. Incorporating more scenarios and datasets to train the model given more computation may also show considerable promise in expanding the current performance.

Additionally, we will conduct specific finetuning of our finetuned models on diverse weather scenarios to further improve performance across different environmental conditions. This weather-specific finetuning will target particular challenges such as snow accumulation, sun glare, and night-time lighting, allowing the system to better adapt to seasonal and temporal variations in driving conditions.

\section{Acknowledgements}
We would like to express our sincere gratitude to the authors of Uncertainty-Aware Adapter: Adapting Segment Anything Model (SAM) for Ambiguous Medical Image Segmentation \cite{jiang2024UATadapt} for their invaluable contributions to the field. Their research and application of uncertainty-aware adapters provided critical insights that significantly influenced our work.

We also deeply appreciate the open-source efforts and datasets made available by the research community, which played a crucial role in facilitating our experimentation and validation. The collaborative spirit of the academic and engineering communities has been instrumental in shaping our study.

\newpage

\bibliographystyle{apalike}
\bibliography{references}

\end{document}